\documentclass[
 two-column
]{ceurart}

\usepackage{listings}
\usepackage{amsmath}
\usepackage{url}
\begin{document}

\copyrightyear{2024}
\copyrightclause{Copyright for this paper by its authors.
  Use permitted under Creative Commons License Attribution 4.0
  International (CC BY 4.0).}

\conference{Forum for Information Retrieval Evaluation, December 12-15, 2024, India}

\title{Prompt Engineering Using GPT for Word-Level Code-Mixed Language Identification in Low-Resource Dravidian Languages}


\author[1]{Aniket Deroy}[%
orcid=0000-0001-7190-5040,
email=roydanik18@kgpian.iitkgp.ac.in,
]
\cormark[1]
\fnmark[1]
\address[1]{IIT Kharagpur,
  Kharagpur, India}

\author[1]{Subhankar Maity}[%
orcid=0009-0001-1358-9534,
email=subhankar.ai@kgpian.iitkgp.ac.in,
]

\cortext[1]{Corresponding author.}

\begin{abstract}
Language Identification (LI) is crucial for various natural language processing tasks, serving as a foundational step in applications such as sentiment analysis, machine translation, and information retrieval. In multilingual societies like India, particularly among the youth engaging on social media, text often exhibits code-mixing, blending local languages with English at different linguistic levels. This phenomenon presents formidable challenges for LI systems, especially when languages intermingle within single words. Dravidian languages, prevalent in southern India, possess rich morphological structures yet suffer from under-representation in digital platforms, leading to the adoption of Roman or hybrid scripts for communication. This paper introduces a prompt based method for a shared task aimed at addressing word-level LI challenges in Dravidian languages. In this work, we leveraged GPT-3.5 Turbo to understand whether the large language models are able to classify words into correct categories correctly. Our findings show that the results on the Kannada dataset consistently outperformed the Tamil dataset across most metrics, indicating a higher accuracy and reliability in identifying and categorizing Kannada language instances. In contrast, the results on the Tamil dataset showed moderate performance, particularly needing improvement across all metrics.
\end{abstract}

\begin{keywords}
  GPT \sep
  Word level identification \sep
  Classification \sep
  Low-resource Languages \sep
  Prompt Engineering 
\end{keywords}

\maketitle

\section{Introduction}

Language Identification (LI)~\cite{jauhiainen2018automatic} is a fundamental task in natural language processing (NLP) that involves determining the language(s) present in a given text. This task is pivotal for numerous applications such as sentiment analysis, machine translation, information retrieval, and natural language understanding. Accurate LI becomes particularly challenging in multilingual societies where texts often exhibit code-mixing, a phenomenon where multiple languages co-occur within the same discourse, ranging from phrases to individual words.

In the context of India, a country renowned for its linguistic diversity~\cite{mandal2019multilingual}, social media platforms reflect a vibrant mix of languages. Among the youth, in particular, there is a prevalent use of code-mixed text that blends local languages from the Dravidian language family with English. Dravidian languages, spoken predominantly in southern India, including languages like Kannada, Tamil, Malayalam, and Tulu, are characterized by rich morphological structures and diverse linguistic features. However, despite their significance, these languages face technological challenges, such as inadequate digital representation and script variations, which complicate language processing tasks like LI.

This paper focuses on addressing the specific challenges of word-level LI in Dravidian languages, leveraging the unique linguistic characteristics and code-mixed nature prevalent in social media and digital communications. We introduce a prompt engineering based method aimed at advancing LI capabilities in these languages by experimenting at different temperature values. By doing so, we aim to contribute to the broader goal of enhancing NLP tools for under-resourced languages, ultimately facilitating more accurate and inclusive language processing technologies.

\begin{figure}[h!]
  \centering
  \includegraphics[width=0.55\linewidth]{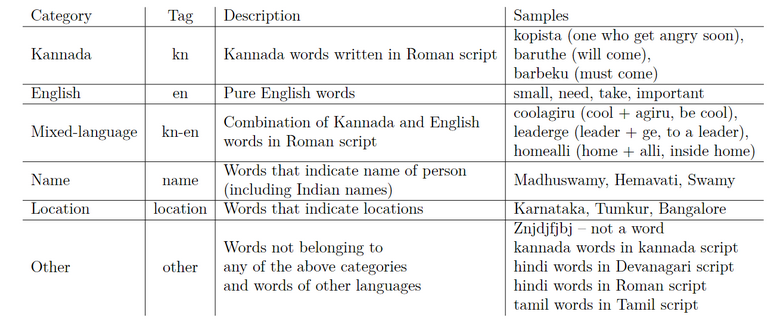}
  \caption{Dataset structure and word categories for the task} \label{fig10}
\end{figure}

An example of the dataset structure and word categories (adapted from-\url{https://sites.google.com/view/coli-dravidian-2024/datasets?authuser=0}) for the task is shown in Figure ~\ref{fig10}.

To the best of our knowledge, there is no work which explores unsupervised approaches for language identification.
In this work, we leveraged GPT-3.5 Turbo \cite{coli7} to understand whether the large language models is able to correctly classify words into correct categories. We experiment with GPT at different temperature values namely 0.7, 0.8, and 0.9.

GPT models are trained on large corpora from the internet, but the availability of high-quality data in Dravidian languages is limited compared to more widely spoken languages like English, Spanish, or Chinese. This means that GPT might not have been exposed to as much diverse or extensive data in these languages. Dravidian languages use distinct scripts (e.g., Tamil script for Tamil, Kannada script for Kannada). Moreover, code-mixing (where Dravidian languages are mixed with English or Hindi, often using the Roman script) is common on social media and informal communications. GPT's ability to handle code-mixed text varies and may not be as robust as its handling of pure English text.

Based on our experiments we observe that for Tamil and Kannada, GPT models have significant room for improvement.

\section{Related Work}

Language Identification (LI)~\cite{jauhiainen2019survey,muthusamy1994automatic,tiedemann2009news} has been a crucial area of research within Natural Language Processing (NLP) due to its foundational role in various applications such as sentiment analysis, machine translation, and information retrieval. Traditional LI approaches~\cite{zampieri2014system,malmasi2015discriminating,king1996labeling,singh2018automatic,zwarts2017proceedings} have primarily focused on monolingual or bilingual sentences, where clear boundaries between languages are assumed. However, these methods often struggle in multilingual and code-mixed environments, especially in regions like India, where linguistic diversity~\cite{tiedemann2003automatic,jauhiainen2017automatic,mandal2015automatic,gamback2016comparing,king2014labeling,molaei2020cross} is high and social media usage reflects complex language practices.

Code-mixing~\cite{zampieri2019predicting,bardaji2012language} presents unique challenges for LI systems. Early research in code-mixing focused on language pairs like English-Spanish or Hindi-English, where code-mixed texts predominantly used Roman scripts. Notable works explored the linguistic features of code-switched texts and highlighted the difficulties in segmenting and identifying languages at the word level. Similarly, Hindi-English code-mixed social media text, emphasizes the necessity for specialized LI models capable of handling intra-word language switches.

Dravidian languages~\cite{chakravarthi2021findings} have been relatively underexplored in the context of LI, primarily due to the scarcity of annotated datasets and the complex morphological characteristics inherent to these languages. Previous efforts have developed initial datasets and models for LI in Dravidian languages; however, these models often fall short in handling code-mixed text, where Roman or hybrid scripts are employed. The Dravidian-CodeMix shared task aimed to address some of these gaps by introducing datasets for Tamil, Malayalam, and Kannada, which included code-mixed instances. Yet, the performance of models on these datasets indicated significant room for improvement, particularly in distinguishing between closely related languages and dialects.

Large Language Models (LLMs)~\cite{radford2019language,raffel2020exploring,devlin2019bert,liu2019roberta} like GPT-3 have shown promise in various NLP tasks, including LI. Previous works have demonstrated the capability of GPT-3 in performing zero-shot and few-shot learning, making it a potentially powerful tool for LI in resource-constrained settings. However, the application of LLMs~\cite{zhao2023survey,vaswani2017attention,gpt2-fine-tuning,zellers2019defending} to code-mixed and morphologically rich languages remains underexplored. Recent studies, have started to explore the use of transformers and pre-trained models for multilingual LI, but the effectiveness of these models in code-mixed Dravidian languages, particularly at the word level, requires further investigation.

Our work builds upon these existing efforts by focusing on a prompt-based method using GPT-3.5 Turbo to address word-level LI challenges in Dravidian languages. Unlike previous approaches, we leverage the linguistic diversity and code-mixed nature of the datasets to enhance the robustness of LI systems in detecting and classifying under-resourced languages. This study contributes to the growing body of research by providing a prompt engineering based method for Kannada, Tamil and evaluating the performance of advanced LLMs in this complex linguistic landscape.

\section{Dataset}

This shared task (adapted from ~\url{https://sites.google.com/view/coli-dravidian-2024/datasets?authuser=0}) consists of four distinct datasets~\cite{hegde2022corpus,lakshmaiah2022coli,hegde2023coli,hegde2023overview,balouchzahi2022overview,10.1145/3632754.3633075,shashirekha2022CoLI,overviewcolidravidian,balouchzahi2022overview,overviewkanglish}:

\begin{enumerate}
    
\item \textbf{Tulu Dataset:}
This dataset is composed of 7,171 code-mixed sentences gathered from YouTube videos. These sentences have been cleaned to remove non-textual elements and transliterated into Roman script. The dataset contains a total of 36,002 words, which are organized into six categories: 'English', 'Kannada', 'Tulu', 'Location', 'Name', and 'Mixed-language'. The dynamic and context-specific nature of mixed-language words presents notable challenges for processing.


\item \textbf{Kannada Dataset:}
This Kannada dataset contains 14,847 tokens in Roman script and is divided into six categories: 'English', 'Kannada', 'Name', 'Mixed-Language', 'Other', and 'Location'. The primary goal of the dataset is to improve techniques for language identification and classification, particularly for Kannada-English code-mixed texts.


\item \textbf{Tamil Dataset:}
The Tamil dataset comprises 17,568 tokens, created using a methodology similar to that employed for the Kannada and Tulu datasets. It is divided into six categories and is designed to facilitate a range of NLP tasks tailored to the Tamil language.


\item \textbf{Malayalam Dataset:}
This dataset consists of 25,035 tokens classified into 7 categories: 'Number', 'Mixed', 'English', 'Location', 'Name', 'sym' (for sentence boundaries), and 'Malayalam'. This dataset offers extensive coverage for NLP tasks and includes the 'Number' category for numerical values, akin to the structure of the other provided datasets.


\end{enumerate}
We participated in shared tasks based on two languages, namely, \textit{Kannada} and \textit{Tamil}.
The test dataset size for Kannada is 2502 words. The test dataset size for Tamil is 2024 words.

\section{Task Definition}
The goal of this task is to classify individual words from a code-mixed text into predefined categories or classes. The words should be classified into the following categories:
\begin{itemize}
    
\item \textbf{\textit{English}:} Words or phrases that are in the English language (e.g., hello, book, run).

\item \textbf{\textit{Dravidian}:} Words or phrases that are in the Kannada language or Tamil language.
    
\item \textbf{\textit{Mixed}:} Words or phrases that mix English, Kannada, or Tamil or combine elements from both languages.
    
\item \textbf{\textit{Name}:} Proper nouns, including names of people, organizations, etc. (e.g., John, Infosys).
    
\item \textbf{\textit{Location}:} Names of places, such as cities, countries, or landmarks (e.g., Bangalore, India, Taj Mahal).
    
\item \textbf{\textit{Symbol}:} Symbols or punctuation marks used in the text (e.g., $*$, $=$, \#, ;).
    
\item \textbf{\textit{Other}:} Words or elements that do not fit into the above categories or are ambiguous.

\end{itemize}

\section{Methodology}
\subsection{Why Prompting?}
Prompting \cite{coli1} to solve a word-level classification problem often arises from the need to accurately identify and categorize individual words within texts that exhibit code-mixing or multilingual content. Next we discuss the reasons why the problem of language identification is tried via prompting through GPT-3.5 Turbo:

\begin{itemize}[-]
   
\item \textbf{Code-Mixing in Texts:} In multilingual societies or digital platforms, texts frequently mix languages, such as local languages with English \cite{coli2}. Understanding which language each word belongs to is essential for applications like sentiment analysis, machine translation, and information retrieval.

\item \textbf{Accuracy in Language Processing:} For effective natural language processing (NLP), identifying the language of each word enhances the accuracy of subsequent tasks \cite{coli3}. It ensures that language-specific models or algorithms are applied correctly.

\item \textbf{Contextual Understanding:} Words in code-mixed texts can change meaning based on the language they are derived from \cite{coli4}. Accurate language identification at the word level aids in preserving context and meaning during NLP tasks.


\item \textbf{Challenges and Innovation:} Word-level classification poses challenges due to the intricacies of code-mixed languages, where words may seamlessly blend multiple languages or scripts \cite{coli6}. Addressing these challenges fosters innovation in NLP methodologies and technologies.
 
\end{itemize}
In summary, prompting to solve word-level classification problems stems from the practical need to accurately handle code-mixed languages and optimize language-specific processing in diverse linguistic contexts.

\subsection{Prompt Engineering-Based Approach}
We used the GPT-3.5 Turbo model via prompting\footnote{\url{https://platform.openai.com/docs/models/gpt-3-5-turbo}} to solve the classification task in Zero-shot mode.
After the prompt is provided to the LLM, the following steps occur internally while generating the output. GPT-3.5 Turbo follows a decoder-only architecture. So based on~\cite{coli7,vaswani2017attention,radford2018improving}, we list these steps, summarizing the prompting approach using GPT-3.5 Turbo. The set of steps~\cite{vaswani2017attention,radford2018improving} for GPT-3.5 Turbo \cite{coli7} is as follows: \\

\textbf{Step 1: Tokenization}

\begin{itemize}
    \item \textbf{Prompt:} \( X = [x_1, x_2, \dots, x_n] \)
    \item The input text (prompt) is first tokenized into smaller units called tokens. These tokens are often subwords or characters, depending on the model's design.
    \item \textbf{Tokenized Input:} \( T = [t_1, t_2, \dots, t_m] \)
\end{itemize}

\textbf{Step 2: Embedding}

\begin{itemize}
    \item Each token is converted into a high-dimensional vector (embedding) using an embedding matrix \( E \).
    \item \textbf{Embedding Matrix:} \( E \in \mathbb{R}^{|V| \times d} \), where \( |V| \) is the size of the vocabulary and \( d \) is the embedding dimension.
    \item \textbf{Embedded Tokens:} \( T_{\text{emb}} = [E(t_1), E(t_2), \dots, E(t_m)] \)
\end{itemize}

\textbf{Step 3: Positional Encoding}

\begin{itemize}
    \item Since the model processes sequences, it adds positional information to the embeddings to capture the order of tokens.
    \item \textbf{Positional Encoding:} \( P(t_i) \)
    \item \textbf{Input to the Model:} \( Z = T_{\text{emb}} + P \)
\end{itemize}

\textbf{Step 4: Attention Mechanism (Transformer Architecture)}

\begin{itemize}
    \item \textbf{Attention Score Calculation:} The model computes attention scores to determine the importance of each token relative to others in the sequence.
    \item \textbf{Attention Formula:}
    \begin{equation}
    \text{Attention}(Q, K, V) = \text{softmax}\left(\frac{QK^T}{\sqrt{d_k}}\right)V
    \end{equation}
    \item where \( Q \) (query), \( K \) (key), and \( V \) (value) are linear transformations of the input \( Z \).
    \item This attention mechanism is applied multiple times through multi-head attention, allowing the model to focus on different parts of the sequence simultaneously.
\end{itemize}

\textbf{Step 5: Feedforward Neural Networks}

\begin{itemize}
    \item The output of the attention mechanism is passed through feedforward neural networks, which apply non-linear transformations.
    \item \textbf{Feedforward Layer:}
    \begin{equation}
    \text{FFN}(x) = \max(0, xW_1 + b_1)W_2 + b_2
    \end{equation}
    \item where \( W_1, W_2 \) are weight matrices and \( b_1, b_2 \) are biases.
\end{itemize}

\textbf{Step 6: Stacking Layers}

\begin{itemize}
    \item Multiple layers of attention and feedforward networks are stacked, each with its own set of parameters. This forms the "deep" in deep learning.
    \item \textbf{Layer Output:}
    \begin{equation}
    H^{(l)} = \text{LayerNorm}(Z^{(l)} + \text{Attention}(Q^{(l)}, K^{(l)}, V^{(l)}))
    \end{equation}
    \begin{equation}
    Z^{(l+1)} = \text{LayerNorm}(H^{(l)} + \text{FFN}(H^{(l)}))
    \end{equation}
\end{itemize}

\textbf{Step 7: Output Generation}

\begin{itemize}
    \item The final output of the stacked layers is a sequence of vectors.
    \item These vectors are projected back into the token space using a softmax layer to predict the next token or word in the sequence.
    \item \textbf{Softmax Function:}
    \begin{equation}
    P(y_i|X) = \frac{\exp(Z_i)}{\sum_{j=1}^{|V|} \exp(Z_j)}
    \end{equation}
    \item where \( Z_i \) is the logit corresponding to token \( i \) in the vocabulary.
    \item The model generates the next token in the sequence based on the probability distribution, and the process repeats until the end of the output sequence is reached.
\end{itemize}

\textbf{Step 8: Decoding}

\begin{itemize}
    \item The predicted tokens are then decoded back into text, forming the final output.
    \item \textbf{Output Text:} \( Y = [y_1, y_2, \dots, y_k] \)
\end{itemize}



\begin{figure}[h!]
  \centering
  \includegraphics[width=0.55\linewidth]{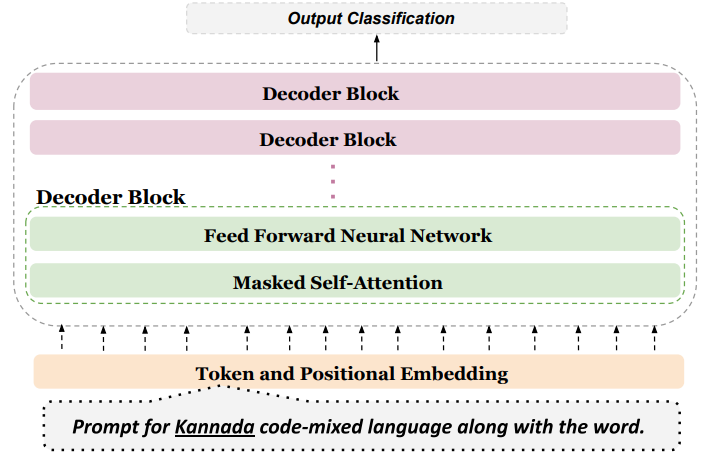}
  \caption{An overview of GPT-3.5 Turbo for Kannada code-mixed language classification.} \label{fig2}
\end{figure}

We used the following prompt for Kannada language for the purpose of classification: "\textit{Please identify which category the word is in English, Kannada, Mixed, Name, Location, Symbol and Other. Please state en, kn, mixed, name, location, sym and other. The word is <Word>}." The figure representing the methodology is shown in Figure ~\ref{fig2}.

\begin{figure}[h!]
  \centering
  \includegraphics[width=0.55\linewidth]{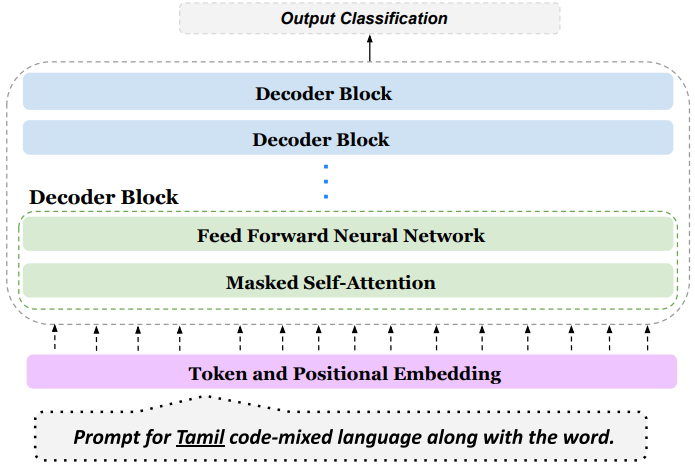}
  \caption{An overview of GPT-3.5 Turbo for Tamil code-mixed language classification.} \label{fig3}
\end{figure}

We used the following prompt for Tamil language for the purpose of classification: "\textit{Please identify which category the word is in English, Tamil, Mixed, Name, Location, Symbol and Other. Please state en, tm, tmen, name, Location, sym and Other. The word is <Word>.}" The figure representing the methodology is shown in Figure ~\ref{fig3}.

Corresponding to the two distinct prompts (for Kannada and Tamil) the two distinct figures are stated (Figure \ref{fig2} and Figure \ref{fig3}).

\section{Results}

\begin{table}[ht]
\centering
\caption{Comparison of various metrics for Word level identification in code-mixed languages in two languages-Tamil, Kannada. The team name is \textbf{TextTitans} and the username is \textbf{roydanik18}. All metric values have been reported in this table.}
\begin{tabular}{|l|c|c|c|c}
\toprule
\toprule
Metric           & Tamil  & Kannada  \\
\midrule
Macro F1               & 0.3312       & 0.4493          \\
Macro Precision        & 0.3259       & 0.5474          \\
Macro Recall           & 0.3657        & 0.4241          \\
Weighted F1      & 0.7022       & 0.6725          \\
Weighted Precision & 0.7559   &    0.7191          \\
Weighted Recall  & 0.6689       & 0.6994          \\
Accuracy  & 0.6689  & 0.6994         \\

\bottomrule
\bottomrule
\end{tabular}

\label{tab:metrics}
\end{table}

Table \ref{tab:metrics} presents metrics comparing the performance of two language identification models, one for Tamil and the other for Kannada.
Our Team ranked in the 10th position for both tasks. Here’s a detailed discussion of each metric.

Next, we discuss the metric values as well as their corresponding analysis to explain the results in the Table ~\ref{tab:metrics}.
For Tamil language, the macro F1 score is 0.3312. This suggests that the model achieves a balanced performance in terms of precision and recall for Tamil language identification. However, it indicates there is room for improvement in correctly identifying both positive and negative instances. For Kannada, the macro F1 score is 0.4493. This score is higher compared to Tamil, indicating a better overall balance between precision and recall for Kannada language identification. The model for Kannada performs better in correctly classifying instances across the dataset.

For Tamil, the macro precision score is 0.3259. For Kannada, the macro precision score is 0.5474. This score indicates a higher accuracy in positive predictions for Kannada compared to Tamil, suggesting better precision in correctly identifying Kannada instances.

For Tamil, the macro recall score is 0.3657. The macro recall score is 0.4241 for Kannada. This score indicates a slightly higher ability to identify Kannada instances correctly compared to Tamil.

For Tamil, the weighted F1 score is 0.7022. This metric considers the F1 score weighted by the number of samples in each class, indicating a solid overall performance for Tamil language identification.
For Kannada, the weighted F1 score is 0.6725. This indicates a slightly lower weighted F1 score compared to Tamil, suggesting a nuanced performance when considering class distribution.

For Tamil, the weighted precision score is 0.7559. This metric reflects the precision of the model when adjusted for the distribution of samples across Tamil language classes. For Kannada, the weighted precision score is 0.7191. This score indicates a slightly lower weighted precision compared to Tamil, reflecting the model’s ability to accurately predict positive instances in Kannada.

For Tamil, the weighted recall score is 0.6689. This metric demonstrates the model's ability to identify all positive instances within the Tamil language classes when considering class distribution.
For Kannada, the weighted recall score is 0.6994. This score indicates a slightly higher ability to correctly identify positive instances within Kannada language classes compared to Tamil.

For Tamil, the accuracy score is 0.6689. This metric measures the overall correctness of the model's predictions for Tamil language identification.
For Kannada, the accuracy score is 0.6994. This indicates a slightly higher overall correctness in predictions for Kannada compared to Tamil.

The metrics highlight differences in performance between the Tamil and Kannada language identification models across various evaluation criteria.
These metrics provide insights into the strengths and areas for improvement in both models, guiding further optimizations and enhancements for accurate language identification tasks in practical applications.

The weighted precision, recall, and f1-scores being higher than the macro precision, recall, and f1-scores shows that dataset likely has an imbalance, with some classes having many more samples than others. The weighted F1 score takes this into account by giving more importance to the performance on larger classes.
The model is performing well on the classes that contribute the most to the overall accuracy. This could mean that it is effectively identifying the majority classes but may struggle with minority classes for both languages.

Weighted precision being higher than weighted recall suggests that the model performs better on the more frequent classes in the dataset. This means it is more effective at correctly identifying positive instances for these majority classes for both datasets.
For Kannada dataset, a higher macro precision than recall may suggest that the model is conservative in its positive predictions, prioritizing accuracy over completeness.
For Tamil dataset, a higher macro recall than precision suggests that while the model is effective at capturing relevant instances, it may not be very reliable in its predictions.

\section{Conclusion}
In this study, we investigated the effectiveness of language identification models for Tamil and Kannada using the advanced capabilities of GPT-3.5 Turbo via prompting. Language identification is a crucial preliminary step in various natural language processing applications, including sentiment analysis, machine translation, and information retrieval. Our research focused on evaluating and comparing the performance of these models across multiple metrics: macro F1 score, macro precision, macro recall, weighted F1, weighted precision, weighted recall, and accuracy. The results reveal notable distinctions between the Tamil and Kannada models. Kannada consistently demonstrated superior performance across most metrics. 
This indicates that the GPT for Kannada effectively identifies and categorizes Kannada language instances with greater accuracy and reliability. Conversely, while the Tamil model exhibited moderate performance, there remains room for improvement, particularly in precision and recall metrics.

The methodology employed in this research leveraged GPT-3.5 Turbo via prompting, harnessing its natural language processing capabilities to handle code-mixed texts and diverse linguistic patterns prevalent in real-world applications. This approach allowed for comprehensive evaluation under varying linguistic contexts, ensuring robustness and applicability in multilingual environments.

Moving forward, further refinements in model training and dataset augmentation could enhance the performance of language identification systems for both Tamil and Kannada. Future research efforts may focus on incorporating additional linguistic features, optimizing model architectures, and expanding datasets to include more diverse linguistic variations and challenges. In conclusion, this study underscores the importance of tailored approaches in language identification, particularly in multilingual settings like India where linguistic diversity is prominent. By advancing the capabilities of language identification models through innovative methodologies such as GPT-3.5 Turbo via prompting, we contribute to the broader goal of improving language processing technologies for diverse and under-resourced languages, fostering more accurate and inclusive natural language understanding systems.
Future work would focus on improving the prompts to improve accuracy on the language identification task.


\bibliography{sample-ceur}

\appendix



\end{document}